\pdfoutput=1

\documentclass[11pt]{article}

\usepackage[final]{acl}

\usepackage{times}
\usepackage{latexsym}

\usepackage[T1]{fontenc}

\usepackage[utf8]{inputenc}

\usepackage{microtype}

\usepackage{inconsolata}

\usepackage{graphicx}

\usepackage{subcaption}

\usepackage{tabularray}

\usepackage{amsmath}

\usepackage[utf8]{inputenc}
\usepackage{geometry}
\usepackage{titlesec}
\usepackage{enumitem}
\usepackage{xcolor}
\usepackage{tcolorbox}

\usepackage{booktabs}
\usepackage{longtable}
\usepackage{array}
\usepackage{caption}
\usepackage{array}
\usepackage{ragged2e}
\usepackage{graphicx}
\usepackage{booktabs}
\usepackage{rotating}

\usepackage{tabularray}
\UseTblrLibrary{booktabs}

\newtcolorbox{dimensionbox}[1]{
    colback=gray!5,
    colframe=gray!50,
    fonttitle=\bfseries\large,
    title=#1,
    arc=3mm,
    boxrule=1pt
}

\title{Differentially-Private Text Rewriting reshapes Linguistic Style}

\author{Stefan Arnold \\ 
Friedrich-Alexander-Universität \\ Erlangen-Nürnberg, Germany \\ 
\texttt{stefan.st.arnold@fau.de}}

\begin{document}
\maketitle
\begin{abstract}

Differential Privacy (DP) for text matured from disjointed word-level substitutions to contiguous sentence-level rewriting by leveraging the generative capacity of language models. While this form of text privatization is best suited for balancing formal privacy guarantees with grammatical coherence, its impact on the register identity of text remains largely unexplored. By conducting a multidimensional stylistic profiling of differentially-private rewriting, we demonstrate that the cost of privacy extends far beyond lexical variation. Specifically, we find that rewriting under privacy constraints induces a systematic functional mutation of the text's communicative signature. This shift is characterized by the severe attrition of interactive markers, contextual references, and complex subordination. By comparing autoregressive paraphrasing against bidirectional substitution across a spectrum of privacy budgets, we observe that both architectures force convergence toward a \textit{non}-involved and \textit{non}-persuasive register. This register-blind sanitization effectively preserves semantic content but structurally homogenizes the nuanced stylistic markers that define human-authored discourse.
\end{abstract}

\section{Introduction}

\textit{Language Models} (LMs) \citep{radford2018improving,brown2020language} have demonstrated a remarkable capacity to capture and reproduce properties of their training data. While this emergent capability underpins their success across a wide range of natural language processing tasks, it has also been shown to expose sensitive information. Prior work has documented that LMs can reveal authorship \citep{song2019auditing} and disclose personally identifiable information \citep{carlini2021extracting,nasr2023scalable}, largely as a consequence of unintended memorization \citep{carlini2019secret}. This raised substantial privacy concerns and led to the development of mechanisms for text privatization.

To mitigate the potential risk of information leakage, researchers have adopted \textit{Differential Privacy} (DP) \citep{dwork2006calibrating}. Originally developed for structured databases, DP formalizes privacy through a rigorous notion of indistinguishability obtained by injecting noise so that the inclusion or exclusion of any individual sample does not substantially affect the outcome distribution.  

Applying DP to text presents unique challenges due to the discrete and highly structured nature of language. \citet{feyisetan2020privacy} addressed this challenge at the word-level by perturbing words in the embedding space \citep{mikolov2013efficient} with semantically proximate substitutions. By framing this process as a randomized response mechanism \citep{warner1965randomized}, this approach provides plausible deniability \citep{bindschaedler2017plausible}. However, by operating on isolated lexical units, this approach tends to degrade grammatical coherence and produce disjointed sentences \citep{arnold2023guiding}. 


To overcome this limitation, the field has transitioned from \textit{disjointed} word-level perturbations toward more \textit{fluent} sentence-level rewriting by leveraging the generative capacity of LMs \citep{mattern2022limits, igamberdiev2023dp, meisenbacher2024dp}. This paradigm shift toward rewriting entire sequences rather than perturbing words in isolation allows for the generation of privatized text that maintains grammatical integrity while satisfying privacy constraints.


While DP rewriting powered by LMs retains semantic content, its impact on \textit{linguistic style} remains insufficiently understood \citep{ccano2025differentially}. This limitation is critical. Since linguistic variation encodes communicative intent, stylistic features determine how meaning is conveyed within a social and functional context \citep{halliday2013halliday}. By shaping distinctions such as a personal opinion and a formal discourse, style governs how texts are perceived \citep{biber1991variation}. Consequently, when rewriting text under privacy constraints, DP mechanisms may not only alter  stylistic fingerprints but fundamentally distort the communicative intent.


\paragraph{Contribution.} We conduct a profiling of stylistic variation of texts obtained from rewriting under constraints of DP. By contrasting autoregressive and bidirectional architectures, we isolate to which extent stylistic distortion arises from the interaction between privacy budget and model design. Specifically, we employ autoregressive rewriting from a fine-tuned LM \citep{mattern2022limits} and bidirectional rewriting from a pre-trained LM \citep{meisenbacher2024dp}, and evaluate them for 67 lexico-grammatical features derived from \citet{biber1991variation} using the \textit{Corpus of Online Registers of English} (CORE) \citep{egbert2015developing}. This corpus comprises 47 diverse registers that range from personal blogs to formal documents. By tracing the degradation of linguistic style down to grammatical features, we can dissect the extent to which DP distorts communicative intent at multiple levels of abstraction. We make three primary contributions showing that DP rewriting does not merely introduce noise at the level of lexemes, but systematically reshapes linguistic style:

\begin{enumerate}
    \item We establish the stylistic preservation by measuring the distributional distance between human and rewritten style features, as a function of the privacy budget. Our results indicate that bidirectional rewriting captures and reproduces human style more effectively than autoregressive paraphrasing across all privacy budgets.  Notably, while the style of the bidirectional approach converges toward the human style as privacy guarantees are relaxed, the autoregressive approach reaches a distinct convergence plateau, indicating persistent stylistic bias. We attribute this to stylistic artifacts stemming from its training history on paraphrasing corpora with limited styles.
    
    \item We pinpoint deviations in feature distributions to show that bidirectional rewriting generally maintains a closer proximity to human baselines than autoregressive rewriting. Both stylometric profiles are characterized by diminished interactive and contextual markers such as personal pronouns and place/time adverbials along with simplified subordinator structure. Our style profiling reveals a sanitization process that under-represents causative (e.g., \textit{because}) and concessive (e.g., \textit{although}) structures but over-generates temporal (e.g., \textit{since}) and adversative subordination (e.g., \textit{whereas}) to maintain a veneer of logical coherence while losing diverse  dependencies.

    \item We project these distributions of features onto the multidimensional framework proposed by \citet{biber1991variation} to interpret the functional transformation of communicative intent. Our analysis reveals that while bidirectional rewriting remains more faithful to the communicative intent compared to autoregressive rewriting, both architectures trigger a fundamental functional shift that often transforms involved and persuasive discourse into sterile and informational reports. This characterizes the register-blind nature of current DP mechanisms which effectively preserve topic and content but fall short of maintaining stylistic identity.
\end{enumerate}


\begin{table*}[!t]
\small
\centering
\caption{Examples from the \texttt{CORE} corpus \cite{egbert2015developing}, categorized according to four \textit{archetypal} dimensions presented by \citet{biber1991variation}. Scores denote the aggregated factor values calculated from the summation of standardized feature frequencies relative to the corpus mean. Representative features are \textit{italicized}.}
\label{tab:biber_archetypes}
\begin{tblr}{
  width = \linewidth,
  colspec = {
    Q[l,m,wd=0.13\linewidth]
    Q[l,m,wd=0.13\linewidth]
    X[l,m]
    Q[c,m,wd=0.08\linewidth]
  },
  row{1} = {font=\bfseries},
  hline{1,Z} = {0.08em},
  hline{2} = {0.05em},
  hline{4,6,8} = {0.03em},
  rowsep = 4pt,
}
Dimension & Pole & Example & Score \\
\SetCell[r=2]{l,m} Focus & Involved & Do \textit{you think not} going to University is a barrier in life, and \textit{you are} unable to be successful? \textit{I used to think this} was a huge barrier. \textit{I didn't} go because \textit{I didn't} have enough money, and \textit{I really don't want to} take student loans, because \textit{I see} [\ldots] & $+41.12$ \\
 & Informational & The \textit{weekly programme} of this \textit{ongoing series} addressing \textit{cutting-edge research} will include \textit{talks} by \textit{invited guest speakers} as well as \textit{original papers} from those teaching across \textit{heritage studies} at \textit{UCL}. [\ldots] & $-19.37$ \\
\SetCell[r=2]{l,m} Discourse & Narrative & An heiress \textit{found} dead in \textit{her} 70 million home in London \textit{had been} due to enter a rehabilitation clinic in California with \textit{her} husband just weeks before, \textit{her} mother \textit{said}. Nancy Kemeny \textit{said} the family \textit{had maintained} ``high hopes'' that [\ldots] & $+17.25$ \\
 & Expository & Here at Taumarunui High School, opportunities for learning abound with a range of subjects to suit even the most exacting study requirements. Courses and subjects offered are as diverse as farming and painting, computing and music, [\ldots] & $-3.55$ \\
\SetCell[r=2]{l,m} References & Situational & [\ldots] The After care report published by Roger Morgan looked at views from 308 care leavers both who had \textit{recently} left care and those \textit{still} in care. The main messages \textit{this year} were that \textit{nearly} half felt they left \textit{too early} and were \textit{not} prepared \textit{well enough}. & $+11.82$ \\
 & Elaborated & The \textit{aim} of this Special \textit{Issue} is to further our \textit{understanding} of the \textit{manner} in \textit{which} [\ldots] impact [\ldots]. Our \textit{intent} is to stimulate critical \textit{debate} and \textit{analyses}. & $-13.25$ \\
\SetCell[r=2]{l,m} Argumentation & Persuasive & [\ldots] You \textit{may think} you \textit{know how to do} everything, but \textit{if} you're a real man, then you \textit{can actually admit} that you probably don't \textit{know how to change} a light bulb. Since every man \textit{should know how to do} this, here are things you \textit{should be} familiar with in order \textit{to be} a man. & $+18.32$ \\
 & Neutral & [\ldots] Quebec is the largest province in Canada by area and borders Ontario, New Brunswick and Newfoundland [\ldots]. The territory of Quebec represents [\ldots] & $-4.33$ \\
\end{tblr}
\end{table*}

\section{Stylometric Analysis}
\label{sec:methodology}

To isolate and measure the stylistic shift introduced by DP rewriting, we contrast the paraphrases of two diametrical model architectures against a human-authored baseline. By leveraging a high-dimensional stylometric framework, we move beyond superficial string-similarity metrics to quantify the functional and communicative distortions that occur under varying privacy constraints.

\subsection{Model Selection and Architecture }

To ensure a controlled comparison of potential architectural impacts under DP constraints, we selected two foundational paradigms representing the primary approaches to sentence-level rewriting: \textsc {DP-Paraphrase} \citep{mattern2022limits} and \textsc{DP-MLM} \citep{meisenbacher2024dp}. We implement \textsc {DP-Paraphrase} utilizing a \texttt{GPT-2} \citep{radford2019language} backbone \textit{fine-tuned} for paraphrasing \citep{dolan2004unsupervised}. This represents an autoregressive approach where privatized text is generated sequentially. \textsc{DP-MLM} employs a \textit{pre-trained} \texttt{RoBERTa} \citep{liu2019roberta} architecture. This represents a bidirectional approach where privatized text is generated by substituting tokens within the full context of the surrounding sentence. We evaluate both approaches across a privacy budget spectrum of $\varepsilon \in \{10, 25, 50, 100, 250\}$, aligning with the benchmarks established by \citet{meisenbacher2024dp} to ensure comparability.

Crucially, both approaches for rewriting rely on temperate sampling \citep{holtzman2020the} to select from the vocabulary distribution. By setting the temperature as a function of the privacy budget, this process operates as an instantiation of the exponential mechanism \citep{mcsherry2007mechanism}. This shared mathematical formulation establishes rigorous experimental control, ensuring that any observed stylistic differences stem from the generative architecture and training history rather than being confounded by discrepancies in the underlying privacy formalization.

We note that we deliberately omit mechanisms at word-level \citep[e.g.,][]{chen2023customized, tian2026stamp} because they frequently fail to preserve basic grammatical coherence \citep{arnold2023guiding}. This limitation confounds stylometric analysis as it is linguistically invalid to meaningfully measure the functional style of a disjointed string.

\subsection{Dataset and Feature Extraction}

We evaluate our selected rewriting mechanisms on a subset of 9,691 documents from the \textit{Corpus of Online Registers of English} (CORE) \citep{egbert2015developing}. Table~\ref{tab:biber_archetypes} provides concrete examples from the CORE corpus. Ranging from personal blogs or discussion forums to legal terms and technical reports, this corpus is uniquely suited for stylometric stress testing due to its unparalleled register diversity found on the open web. 

To ensure linguistic interpretability, we eschew opaque style embeddings \citep{rivera2021learning, patel2023learning} in favor of a curated set of lexico-grammatical features \citep{biber1995dimensions}. While style embeddings proved effective for authorship attribution, they aggregate stylistic variance into often uninterpretable high-dimensional representations. Such representations obscure the specific grammatical shifts induced by DP rewriting, making it impossible to determine whether a shift in the embedding space stems from lexical distortion, syntactic simplification, or a loss of grammatical markers.

In contrast, the framework compiled by \citet{biber1995dimensions} enables the precise quantification of 67 distinct linguistic features, organized into 16 grammatical and functional categories. This comprehensive set spans a wide range of linguistic aspects, including \textit{tense and aspect markers}, \textit{place and time adverbials}, \textit{nominal forms}, \textit{passive voice}, \textit{modality words}, \textit{negation types}, and \textit{subordination features}. Table~\ref{tab:biber_features}, deferred to the Appendix, provides a full inventory of these grammatical features and functional categories. By tracking the frequencies of these grammatical constructs, we can pinpoint exactly what private rewriting fails to preserve, decomposing an abstract measure of stylistic drift into a concrete inventory of grammatical shifts.

\section{Style Variance}

Aimed at dissecting the impact of private rewriting on linguistic style, we adopt a hierarchical analysis across three levels of granularity.

\subsection{Stylistic Fidelity}

To characterize the stylistic shift induced by DP rewriting, we quantify the divergence between original and privatized text using \textit{Burrows' Delta} \citep{burrows2002delta}. This measure captures the distance between standardized relative frequencies of linguistic items and serves as a proxy for the stylistic fingerprint of a text. A low value indicates close adherence to style properties, whereas a high value reflects increased stylistic deviation.

\begin{figure}[!thb]
    \includegraphics[width=0.45\textwidth]{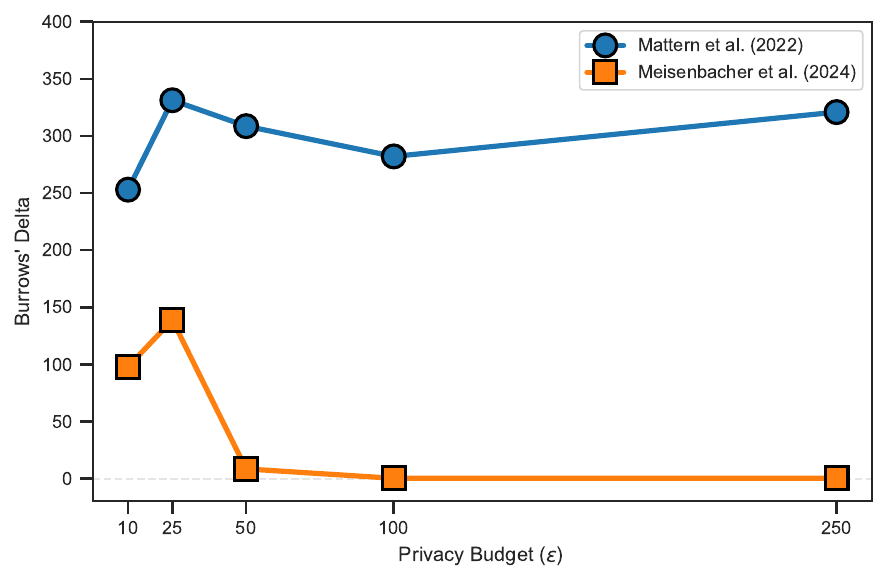}
    \caption{Stylistic deviation of privatized text compared to the human-authored text, measured using \textit{Burrows' Delta}. Both approaches exhibit erratic behavior under tight privacy. \textsc{DP-MLM} consistently demonstrates significantly lower deviation across the entire privacy spectrum and converges toward the human style. \textsc{DP-Paraphrase} reaches a distinct convergence plateau, failing to recover the human style.}
    \label{fig:linguistic_delta}
\end{figure}

Figure~\ref{fig:linguistic_delta} presents the relationship between the privacy budget and stylistic preservation for both autoregressive and bidirectional rewriting, revealing a significant architectural disparity. Across the entire spectrum of privacy budgets, bidirectional privatization enables \textsc{DP-MLM} to better retain distributional properties than the autoregressive privatization process. This is reflected by consistently lower delta values for \textsc{DP-MLM} compared to \textsc{DP-Paraphrase}, indicating a systematically higher degree of stylistic preservation.


\begin{figure*}[!thb]
    \centering
    \includegraphics[width=0.9\textwidth]{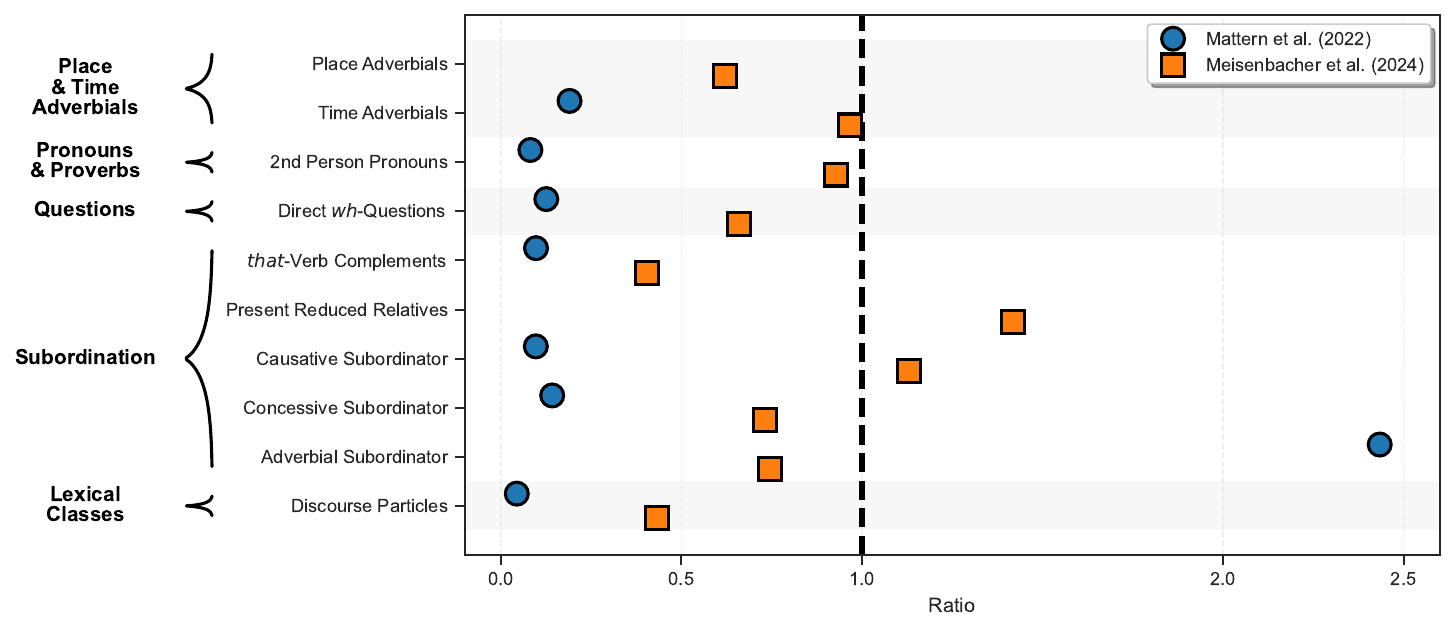}
    \caption{Relative usage ratios of key lexico-grammatical features in privatized text compared to the human-authored \texttt{CORE} baseline (dashed line). Ratios below unity indicate under-representation, while those above signify over-generation. Across all of the most deviating features, \textsc{DP-MLM} \cite{meisenbacher2024dp} maintains usage significantly closer to the human baseline than \textsc{DP-Paraphrase} \cite{mattern2022limits}.
    }
    \label{fig:linguistic_profile}
\end{figure*}

Contrary to a monotonic decay in stylistic distortion given the privacy budgets, we observe that the relationship between the privacy budget and stylistic distance is not strictly monotonic. In the regime of tight privacy, both techniques exhibit fluctuations, showing a slight rise in delta followed by a subsequent dip. This pattern reflects the interaction between the noise injection of the exponential mechanism and the sampling distribution, where extremely high noise levels can occasionally lead to erratic stylistic texts. As privacy constraints are progressively relaxed, both architectures diverge markedly in their convergence behavior. While \textsc{DP-MLM} demonstrates a trajectory toward the original stylistic signature, \textsc{DP-Paraphrase} reaches a distinct plateau and fails to recover the original style even at high privacy budgets.

The persistent divergence in stylistic convergence exhibited by \textsc{DP-Paraphrase} points to an underlying architectural limitation. A plausible explanation for this lack of convergence resides in its training history. Specifically, \textsc{DP-Paraphrase} relies on a \texttt{GPT-2} backbone \citep{radford2019language} fine-tuned for paraphrasing, thereby inheriting the stylistic regularities of the paraphrase corpus \citep{dolan2004unsupervised}. Rather than gradually aligning with the stylistic signature as noise diminishes, the model instead converges toward the paraphrasing artifacts, even with almost no privacy guarantee. This dynamic effectively traps the model within its paraphrasing style. Collectively, these findings establish that stylistic degradation is not solely governed by the privacy budget, but is strongly mediated by architectural priors and training history.

\subsection{Feature Deviations}

To elucidate the linguistic drivers of stylistic divergence, we examine the specific lexico-grammatical features that exhibit the highest degree of deviation from human-authored style. This effectively transforms an abstract measure of stylistic decline into a concrete inventory of linguistic drift. We focus on the ten features exhibiting the largest deviations. By isolating these most deviating features, we can identify which grammatical markers are systematically suppressed or amplified by the privatized models. We quantify feature shifts using the usage ratio $R = f_{\text{model}} / f_{\text{human}}$, where $R = 1$ denotes a perfect preservation of human-like frequency, $R < 1$ indicates under-representation, and $R > 1$ signifies over-generation. This formulation allows for a direct comparison of functional feature usage between original and privatized text.

Figure~\ref{fig:linguistic_profile} reveals a pronounced asymmetry. Most of the features that deviate are substantially underrepresented. We notice a substantial loss of contextual and interactive markers, where \textit{place adverbials}, \textit{time adverbials}, \textit{personal pronouns}, and \textit{direct questions} all exhibit ratios well below unity. This trend suggests that DP rewriting  exerts a sanitizing effect on text, stripping away the specific spatial, temporal, and interpersonal anchors that characterize original communicative intent.  

A similar trend emerges for complex syntactic constructions. \textsc{DP-Paraphrase} flattens subordinators associated with causation and concession and replaces them with a narrow set of temporal and adversative subordinators to maintain logical coherence. Compared to causative and concessive subordinators, \textsc{DP-Paraphrase} over-generates temporal and adversative subordinators with usage frequencies approaching 2.5 times the human baseline, representing the sole feature that is significantly over-generated. In contrast, \textsc{DP-MLM} trades concessive, temporal and adversative subordinators with causative sentence structures. This simplification of syntactic variance suggests that, under privacy constraints, models default to limited repertoire of sentence constructions.

Despite notable deviation from human usage, \textsc{DP-MLM} largely maintains feature usage within a bounded range. For almost all deviating features, its ratios remain approximately within a $\pm 0.5$ interval around the human baseline, indicating that stylistic variation remains within a tolerable margin. This pattern implies that bidirectional substitution preserves a substantial degree of stylistic fidelity, maintaining nuanced stylistic characteristics despite the constraints imposed by the privacy mechanism. In contrast, \textsc{DP-Paraphrase} frequently exceeds the range of tolerable margin. This architectural comparison reinforces that \textsc{DP-MLM} is more stylistically robust than \textsc{DP-Paraphrase}, which tends to prioritize semantic equivalence over the preservation of stylistic properties.

\begin{figure*}[!thb]
    \centering
    \begin{subfigure}[b]{0.23\textwidth}
        \centering
        \includegraphics[width=\textwidth]{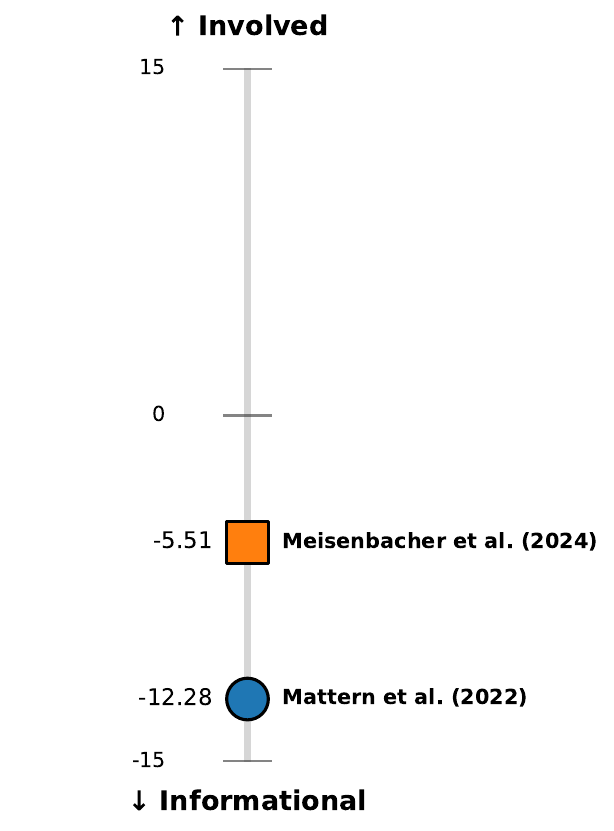}
        \caption{Production}
        \label{fig:spectrum_production}
    \end{subfigure}
    \hfill
    \begin{subfigure}[b]{0.23\textwidth}
        \centering
        \includegraphics[width=\textwidth]{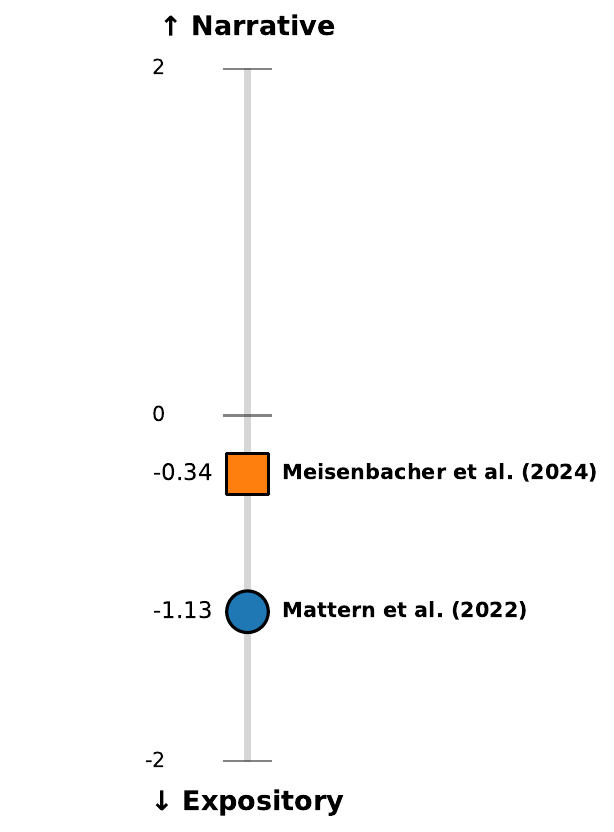}
        \caption{Discourse}
        \label{fig:spectrum_discourse}
    \end{subfigure}
    \hfill
    \begin{subfigure}[b]{0.23\textwidth}
        \centering
        \includegraphics[width=\textwidth]{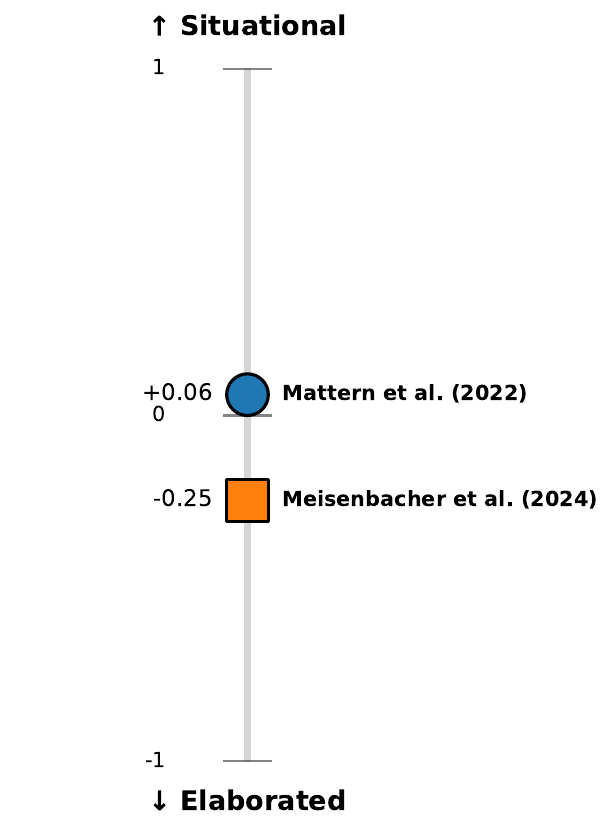}
        \caption{References}
        \label{fig:spectrum_reference}
    \end{subfigure}
    \hfill
    \begin{subfigure}[b]{0.23\textwidth}
        \centering
        \includegraphics[width=\textwidth]{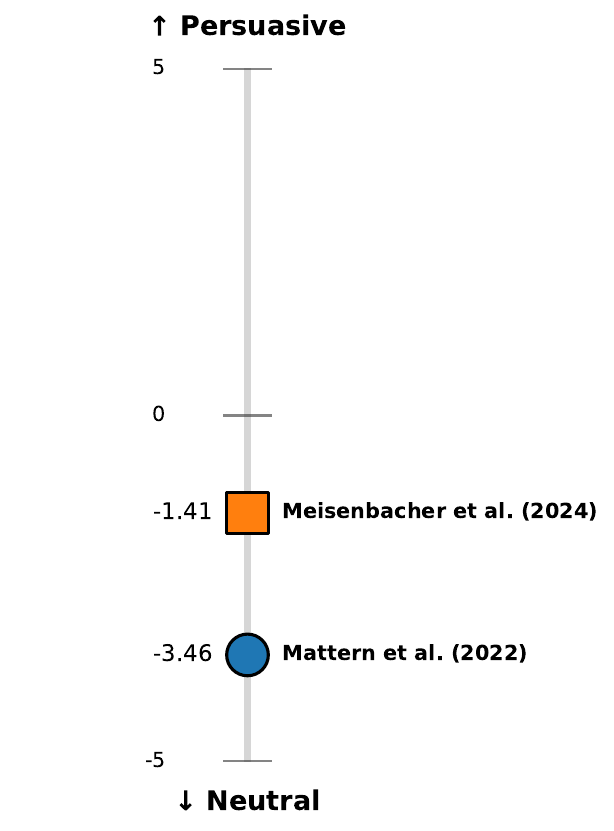}
        \caption{Argumentation}
        \label{fig:spectrum_argument}
    \end{subfigure}
    \caption{Functional deviation of privatized text along the four canonical dimensions interpreted by \citet{biber1991variation}, where the zero-axis represents the human-authored communicative intent. Positive and negative poles correspond to distinct functional registers. Across most dimensions, \textsc{DP-MLM} \cite{meisenbacher2024dp} clusters significantly closer to the original communicative intent than \textsc{DP-Paraphrase} \cite{mattern2022limits}. However, both approaches tend to favor informational, expository, and neutral modes of communication.}
    \label{fig:linguistic_spectrum}
\end{figure*}

\subsection{Functional Shifts}

While lexico-grammatical usage ratios identified the specific linguistic casualties, they do not inherently explain how these isolated changes aggregate to alter the broader communicative intent of the text. To interpret how grammatical variation affects communicative intent, we project the stylistic profiles onto the multidimensional factor analysis introduced by \citet{biber1991variation}. Building on the assumption that cooccurrence patterns reflect underlying communicative functions, factor interpretations dictate which features contribute to the positive or negative pole of a given dimension. We adopt the functional dimensions established by \citet{biber1991variation} and reuse their canonical groupings of positively and negatively associated features — rather than deriving factor loadings strictly from \texttt{CORE}. These factor loadings were derived from extensive and highly representative corpora spanning a wide array of spoken and written English registers, rendering these dimensions as a robust and widely accepted standard in sociolinguistics. By reusing the established positive and negative feature groupings and projecting our scores onto his canonical dimensions, we anchor observed stylistic shifts within this validated framework.

We calculate the dimension score $D_k$ for each text by summing the standardized frequencies of features with positive loadings $z^{(+)}$ and subtracting those with negative loadings $z^{(-)}$:

\[
D_k = \sum z_{i}^{(+)} - \sum z_{j}^{(-)}.
\]

This procedure yields interpretable scores that reflect shifts in communicative function along established axes of variation. We center the human-authored baseline at zero, enabling comparison of communicative shifts induced by privatization.

Figure~\ref{fig:linguistic_spectrum} illustrates the relative deviation of the privatized text from the original communicative intent across the four canonical dimensions. Across all dimensions, privatized texts exhibit a systematic functional drift away from the distinct functional norms of the original text.

\paragraph{Involved vs. Informational Production:} This dimension distinguishes between interactive, affective discourse and abstract, information-dense prose. We observe the most pronounced shift along this dimension. Relative to human discourse, \textsc{DP-Paraphrase} exhibits a marked drift toward an informational style, signaling a systematic loss of interactive and affective markers. This shift manifests as a more sterile, impersonal mode of expression, driven by the suppression of personal pronouns and direct questions. While \textsc{DP-MLM} likewise strips away the involved character of a text, it maintains substantially higher production fidelity, with deviations of roughly half the magnitude observed for \textsc{DP-Paraphrase}. Overall, this transformation recasts dynamic, person-oriented discourse as impersonal, informational reporting.

\paragraph{Narrative vs. Expository Discourse:} This dimension separates narrative discourse, characterized by past-tense construction, from expository styles marked by present-tense usage. \textsc{DP-MLM} remains closely aligned with the narrative degree of human discourse, suggesting the perseveration of discourse structure, whereas \textsc{DP-Paraphrase} drops sharply into an expository style. This pattern is reflected by structural erosion of temporal sequencing and storytelling devices. Consequently, the rewriting process homogenizes narrative arcs, effectively converting narrative storytelling into a series of detached expository utterances.

\paragraph{Situational vs. Elaborated Reference:} This dimension contrasts explicit reference, typically realized through relative clauses, with references that depend heavily on the immediate temporal or physical context. Whereas elaborated reference renders a text autonomous and interpretable without external cues, situational reference is characterized by a high degree of deixis, anchoring communicative intent in expressions such as \textit{here}, \textit{there}, and \textit{now}. \textsc{DP-Paraphrase} largely maintains the level of referential density. \textsc{DP-MLM} elaborates the reference structure by replacing deictic placeholders with more context-independent descriptors. This divergence underscores how architectural constraints shape distinct referential strategies.

\paragraph{Persuasive vs. Neutral Argumentation:} This dimension captures the extent to which a human explicitly marks a point of view through the use of infinitives, modals, and suasive verbs. \textsc{DP-Paraphrase} departs markedly from the human expression of persuasion, exhibiting a significantly more neutral tone. This reduction is mirrored by a systematic attrition of the rhetorical and persuasive force, effectively neutralizing the original argumentative intent. In contrast, this functional shift is far less pronounced for \textsc{DP-MLM}. This divergence suggests that bidirectional substitution is more robust in preserving argumentative structure, whereas autoregressive paraphrasing tends to flatten subjective stances into a more neutral register.

The dimensional analysis reveals that DP rewriting induces systematic transformations to the communicative signature of text. Although both architectures demonstrated to successfully preserve semantic content, \textsc{DP-Paraphrase} and \textsc{DP-MLM} fall short of maintaining the nuanced stylistic variances characteristic of human-authored registers. Owing to its reliance on autoregressive paraphrasing, \textsc{DP-Paraphrase} tends to structurally mutate the functional identity, whereas \textsc{DP-MLM}, through bidirectional substitution, remains comparatively more faithful. Nevertheless, both processes manifest into a forced convergence toward a less interactive and less persuasive mode of communication, confirming that formal privacy extends beyond surface variation to fundamentally reshape the text's communicative function.

\section{Related Work}

Since the inception of DP for text at word-level \citep{feyisetan2020privacy}, significant progress has been made in enhancing privacy guarantees \citep{xu2021density} and task utility \citep{carvalho2021tem}. Recent advancements have further optimized this trade-off through mapping schemes \citep{yue2021differential, chen2023customized} and selective mechanisms \citep{tian2026stamp} that distribute privacy budgets more intelligently. Parallel to these refinements, the field has transitioned toward generating human-readable paraphrases at sentence-level. Early efforts utilized variational autoencoders to perturb latent vectors rather than discrete tokens \citep{weggenmann2022dp}, while more recent work takes advantage of the vocabulary space of conditional \citep{igamberdiev2023dp}, causal \citep{mattern2022limits, utpala2023locally}, and masked \citep{meisenbacher2024dp} language models.

We extend existing analyses into the language quality of differentially-private text. \citet{mattern2022limits} and \citet{arnold2023guiding} examine the retention of morphosyntactic integrity under word-level substitution, whereas \citet{ccano2025differentially} assess text at sentence-level using surface-oriented metrics such as lexical diversity and grammatical correctness. We shift the analytical focus from surface measures to the preservation of stylistic and functional identity in differentially-private text.



\section{Conclusion}

The evolution of differentially-private text rewriting has successfully bridged the gap between formal privacy and grammatical fluency, yet this technical maturation has come at a significant cost to the functional identity of the text. Through a stylometric profiling of lexico-grammatical features and their projection onto broader dimensions of communicative intent, we demonstrate that privacy-constrained rewriting induces systematic stylistic homogenization which is characterized by stripping away the interactive and persuasive markers that anchor human-authored intent.

Beyond this functional shift of communicative intent, our analysis further reveals a critical architectural disparity in how models navigate the privacy-utility trade-off. While the bidirectional substitution of \textsc{DP-MLM} maintains a trajectory of alignment with human stylistic norms as privacy constraints are relaxed, the autoregressive generation of \textsc{DP-Paraphrase} reaches a distinct convergence plateau. This persistent deviation, even without formal privacy guarantees, suggests that the model remains trapped by the stylistic priors of its training history, prioritizing formulaic paraphrasing over the nuanced preservation of grammatical and functional integrity. Advances in DP rewriting depend on moving beyond fluency as the primary objective and addressing stylistic erosion, so that privatized text retains both formal guarantees and the diverse functional integrity of discourse.

\paragraph{Limitations.} We note that the primary limitation of this stylometric analysis lies in the evaluation of linguistic style in isolation from adversarial authorship attribution \citep{huang2024large}. This limitation stems from a fundamental entanglement of register and identity: linguistic variation operates along a continuum from functional registers (e.g., persuasive opinion) to the idiosyncratic fingerprint unique to an author (e.g., punctuation densities). Although demonstrating that DP rewriting sanitizes the communicative intent, inducing a convergence toward information-dense, neutral prose, it remains unclear whether this sanitization is a necessary prerequisite for thwarting deanonymization. 

We therefore plan to extend this line of linguistic inspection by assessing the \textit{Pareto frontier} between the successful retention of register-level functional identity and author-level privacy risks. 

\bibliography{custom}

\appendix
\onecolumn
\section{Appendix}

\begin{longtblr}[
  caption = {Lexico-grammatical features, adapted from \citep{biber1991variation}, describing stylistic variance in English and organized based on grammatical and functional categories. Examples \textit{italicized}.},
  label = {tab:biber_features},
]{
  width = \linewidth,
  colspec = { Q[r, 0.8cm] X[l] },
  rowhead = 0,
  cells = {font=\small},
  hline{1,Z} = {0.08em}, 
  cell{1,5,8,16,18,22,25,28,47,52,55,63,67,72,78,81}{1} = {c=2}{l, font=\bfseries\small}, 
  rowsep = 0.6pt, 
  row{1,5,8,16,18,22,25,28,47,52,55,63,67,72,78,81} = {abovesep=2pt},
}
A. Tense and aspect markers & \\
1 & Past tense (e.g., \textit{walked, saw}) \\
2 & Perfect aspect (e.g., \textit{walked, seen}) \\
3 & Present tense (e.g., \textit{walks, sees}) \\

B. Place and time adverbials & \\
4 & Place adverbials (e.g., \textit{above, beside}) \\
5 & Time adverbials (e.g., \textit{early, soon}) \\

C. Pronouns and proverbs & \\
6 & First-person pronouns (e.g., \textit{I, we, us}) \\
7 & Second-person pronouns (e.g., \textit{you, yours}) \\
8 & Third-person personal pronouns (e.g., \textit{he, she}, excluding \textit{it}) \\
9 & Pronoun \textit{it} \\
10 & Demonstrative pronouns (\textit{that, this, these, those} as pronouns) \\
11 & Indefinite pronouns (e.g., \textit{anybody, nothing, someone}) \\
12 & Proverb \textit{do}  \\

D. Questions & \\
13 & Direct WH questions (e.g., \textit{What did he see?}) \\

E. Nominal forms & \\
14 & Nominalizations (ending in \textit{-tion, -ment, -ness, -ity}) \\
15 & Gerunds (participial forms functioning as nouns) \\
16 & Total other nouns (e.g., \textit{house, dog, idea}) \\

F. Passives & \\
17 & Agentless passives (e.g., \textit{the work was done}) \\
18 & \textit{by}-passives (e.g., \textit{the work was done by ...}) \\

G. Stative forms & \\
19 & \textit{be} as main verb (e.g., \textit{he is happy}) \\
20 & Existential \textit{there} (e.g., \textit{there is a chance}) \\

H. Subordination features & \\
21 & \textit{that} verb complements (e.g., \textit{I said that he went}) \\
22 & \textit{that} adjective complements (e.g., \textit{I'm glad that you like it}) \\
23 & WH-clauses (e.g., \textit{I believed what he told me}) \\
24 & Infinitives (e.g., \textit{to walk, to see}) \\
25 & Present participial adverbial clauses (e.g., \textit{Smiling, Joe left.}) \\
26 & Past participial adverbial clauses (e.g., \textit{Built well, the house would stand for fifty years}) \\
27 & Past participial postnominal clauses (e.g., \textit{the solution produced by this process}) \\
28 & Present participial postnominal clauses (e.g., \textit{The event causing this decline was ...}) \\
29 & \textit{that} relative clauses on subject position (e.g., \textit{the dog that bit me}) \\
30 & \textit{that} relative clauses on object position (e.g., \textit{the dog that I saw}) \\
31 & WH relatives on subject position (e.g., \textit{the man who likes popcorn}) \\
32 & WH relatives on object position (e.g., \textit{the man who Sally likes}) \\
33 & Pied-piping relative clauses (e.g., \textit{the manner in which he was told}) \\
34 & Sentence relatives (e.g., \textit{Bob likes fried mangoes, which is most disgusting.}) \\
35 & Causative adverbial subordinator (\textit{because}) \\
36 & Concessive adverbial subordinators (\textit{although, though}) \\
37 & Conditional adverbial subordinators (\textit{if, unless}) \\
38 & Other adverbial subordinators (\textit{since, while, whereas}) \\

I. Prepositional phrases, adjectives, and adverbs & \\
39 & Total prepositional phrases (e.g., \textit{in the garden, at the office}) \\
40 & Attributive adjectives (e.g., \textit{the big horse}) \\
41 & Predicative adjectives (e.g., \textit{the horse is big}) \\
42 & Total adverbs (e.g., \textit{quickly, very, happily}) \\

J. Lexical specificity & \\
43 & Type-token ratio (i.e., \textit{vocabulary diversity}) \\
44 & Mean word length (i.e., \textit{word complexity}) \\

K. Lexical classes & \\
45 & Conjuncts (e.g., \textit{consequently, furthermore, however}) \\
46 & Downtoners (e.g., \textit{barely, nearly, slightly}) \\
47 & Hedges (e.g., \textit{at about, something like, almost}) \\
48 & Amplifiers (e.g., \textit{absolutely, extremely, perfectly}) \\
49 & Emphatics (e.g., \textit{a lot, for sure, really}) \\
50 & Discourse particles (e.g., sentence-initial \textit{well, now, anyway}) \\
51 & Demonstratives (e.g., \textit{this, that, these, those}) \\

L. Modals & \\
52 & Possibility modals (e.g., \textit{can, may, might, could}) \\
53 & Necessity modals (e.g., \textit{ought, should, must}) \\
54 & Predictive modals (e.g., \textit{will, would, shall}) \\

M. Verb classes & \\
55 & Public verbs (e.g., \textit{assert, declare, mention}) \\
56 & Private verbs (e.g., \textit{assume, believe, doubt, know}) \\
57 & Suasive verbs (e.g., \textit{command, insist, propose}) \\
58 & Speculative verbs (e.g., \textit{seems, appear}) \\

N. Reduced forms and dispreferred structures & \\
59 & Contractions (e.g., \textit{don't, it's, won't}) \\
60 & Subordinator \textit{that} deletion (e.g., \textit{I think [that] he went}) \\
61 & Stranded prepositions (e.g., \textit{the candidate that I was thinking of}) \\
62 & Split infinitives (e.g., \textit{He wants to convincingly prove that...}) \\
63 & Split auxiliaries (e.g., \textit{They were apparently shown to ...}) \\

O. Coordination & \\
64 & Phrasal coordination (e.g., cats \textit{and} dogs; hot \textit{and} cold) \\
65 & Independent clause co-ordination (e.g., clause-initial \textit{and}) \\

P. Negation & \\
66 & Synthetic negation (e.g., \textit{No answer is good enough}) \\
67 & Analytic negation (e.g., \textit{That's not likely}) \\
\end{longtblr}
\end{document}